\documentclass{article}

\usepackage{multicol}
\usepackage{arxiv}

\usepackage[utf8]{inputenc} % allow utf-8 input
\usepackage[T1]{fontenc}    % use 8-bit T1 fonts
\usepackage{hyperref}       % hyperlinks
\usepackage{url}            % simple URL typesetting
\usepackage{booktabs}       % professional-quality tables
\usepackage{amsfonts}       % blackboard math symbols
\usepackage{nicefrac}       % compact symbols for 1/2, etc.
\usepackage{microtype}      % microtypography
\usepackage{lipsum}
\usepackage{cite}
\usepackage{amsmath,amssymb,amsfonts}
\usepackage{algorithmic}
\usepackage{graphicx}
\usepackage{textcomp}
\usepackage{comment}
\usepackage{braket}
\usepackage{physics}
\usepackage[ruled,vlined]{algorithm2e}
\usepackage[fleqn]{nccmath}
\usepackage{mathbbol}
\usepackage{subfigure}
\newcommand{\RN}[1]{%
  \textup{\uppercase\expandafter{\romannumeral#1}}%
}

\title{A Framework for Recognizing and Estimating Human Concentration Levels}

\author{
 Woodo Lee \\
   Department of Physics\\
  Korea University\\
  Seoul, Republic of Korea, 02481 \\
  \texttt{woodolee@korea.ac.kr} \\
   \And
   
 Jakyung Koo \\
    Department of Computer Science and Engineering\\
  Korea University\\
  Seoul, Republic of Korea, 02481 \\
\texttt{lawkelvin33@korea.ac.kr} \\
\And

 Nokyung Park \\
    Department of Computer Science and Engineering\\
  Korea University\\
  Seoul, Republic of Korea, 02481 \\
\texttt{noparkee@korea.ac.kr} \\
\And

 Pilgu Kang \\
    Department of Computer Science and Engineering\\
  Korea University\\
  Seoul, Republic of Korea, 02481 \\
\texttt{rkd903@korea.ac.kr} \\
\And
 Jeakwon Shim \thanks{Corresponding author.} \\
    Department of Computer Science and Engineering\\
  Korea University\\
  Seoul, Republic of Korea, 02481 \\
  \texttt{jaekwoun.shim@gmail.com}
}

\begin{document}
\maketitle

\begin{abstract}
One of the major tasks in online education is to estimate the concentration levels of each student.
Previous studies have a limitation of classifying the levels using discrete states only.
The purpose of this paper is to estimate the subtle levels as specified states by using the minimum amount of body movement data.
This is done by a framework composed of a Deep Neural Network and Kalman Filter. 
Using this framework, we successfully extracted the concentration levels, which can be used to aid lecturers and expand to other areas.

\end{abstract}


% \begin{multicols}{2}
\section{Introduction}
There have been many attempts to measure students' concentration levels using various methods such as taking skin temperature \cite{b1}, recognizing visual attention, and detecting Electroencephalogram (EEG) signals \cite{b2, b3, b4}.
However, these attempts lack in detail since their concentration levels were classified as discrete states.
Here, we develop a new framework (DNN-K), named after its architecture, Deep Neural Network (DNN) and Kalman Filter (KF) \cite{b5} to improve these limitations.
DNN-K defines the concentration levels as the probability of "high concentration," which is derived from the DNN, and suggests that the standard deviations of designated points are a core factor in measuring these probabilities.
KF is also implemented to DNN-K to estimate the concentration levels over measuring.

\section{Motivation}
\begin{figure}
    \centering
    \includegraphics[width=0.7\columnwidth]{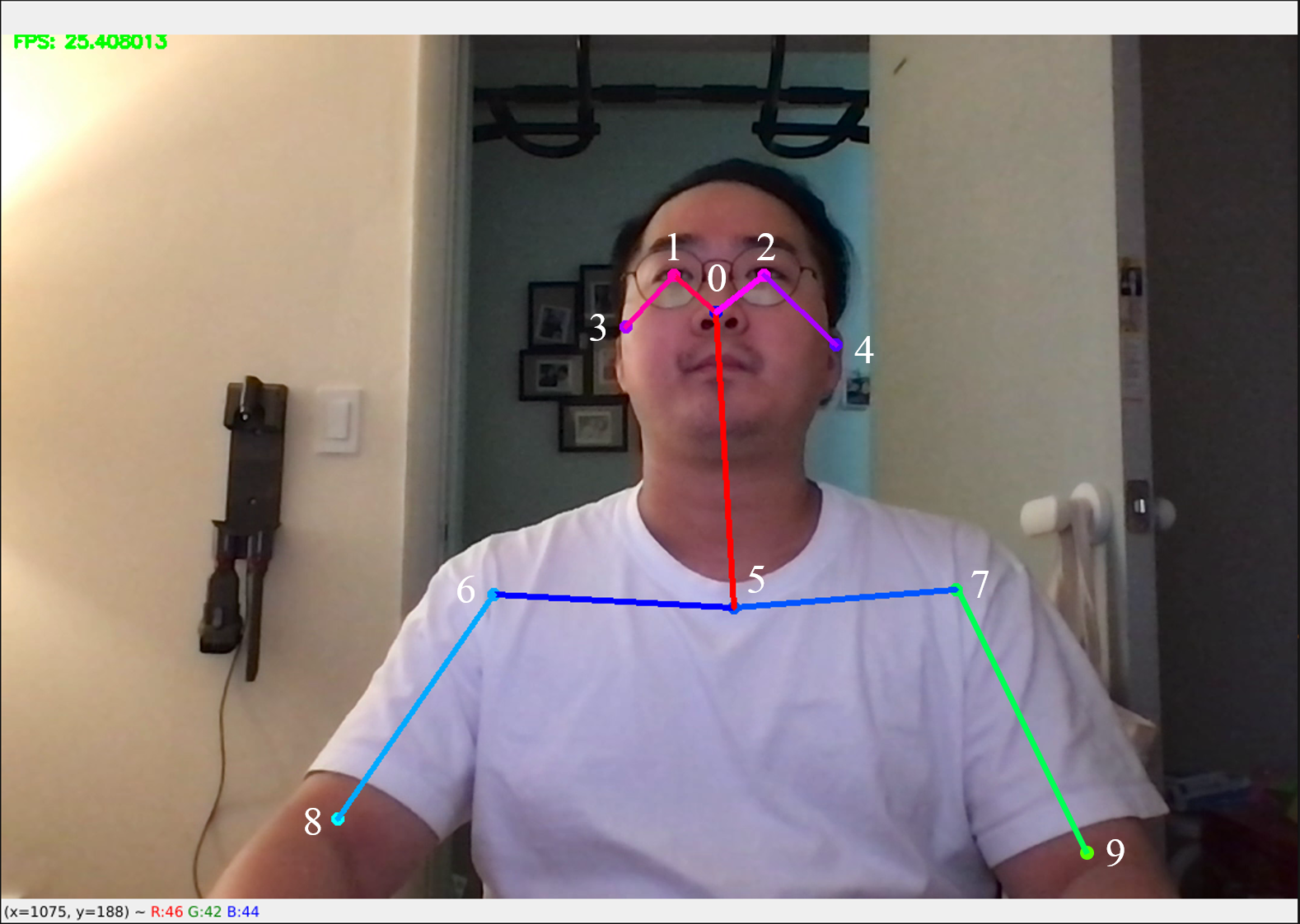}
    \caption{
    The measured points by OpenPose are shown.
    The middle and upper body are being measured with ten points respectively.
    }
    \label{fig:wdlee}
\end{figure}

\begin{figure}
    \centering
    \subfigure[]{
        \includegraphics[width=0.7\columnwidth]{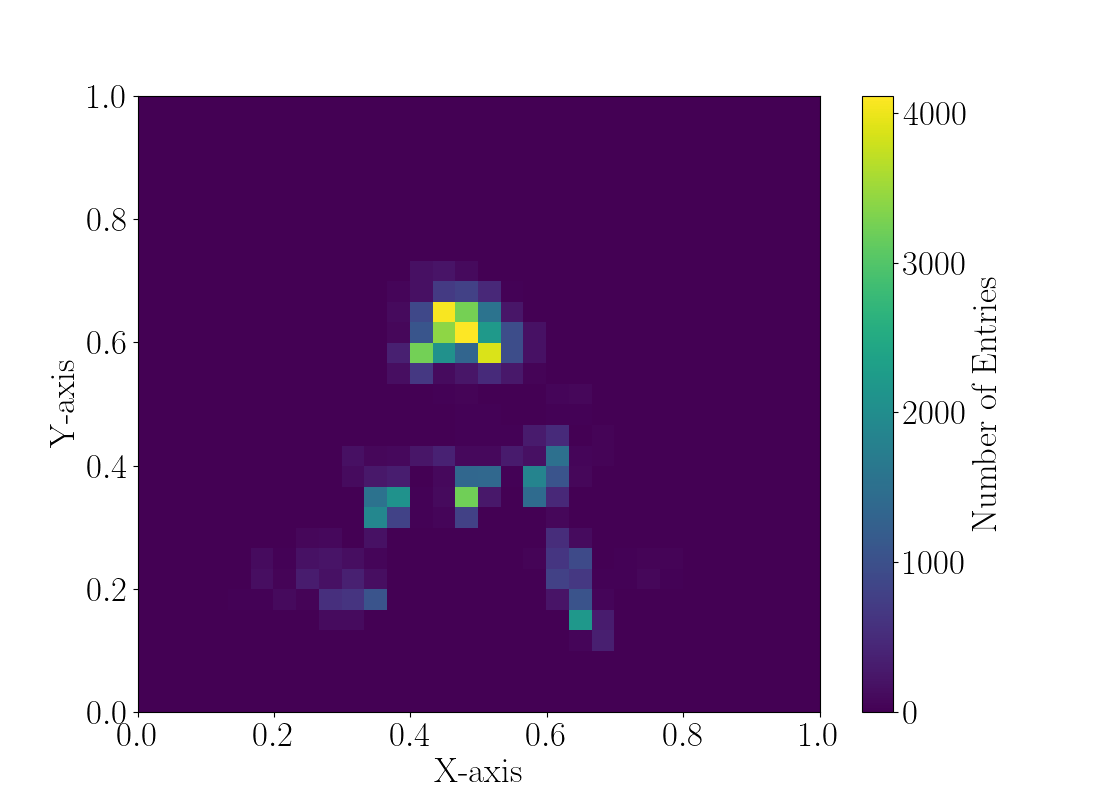}
    }
    \subfigure[]{
        \includegraphics[width=0.7\columnwidth]{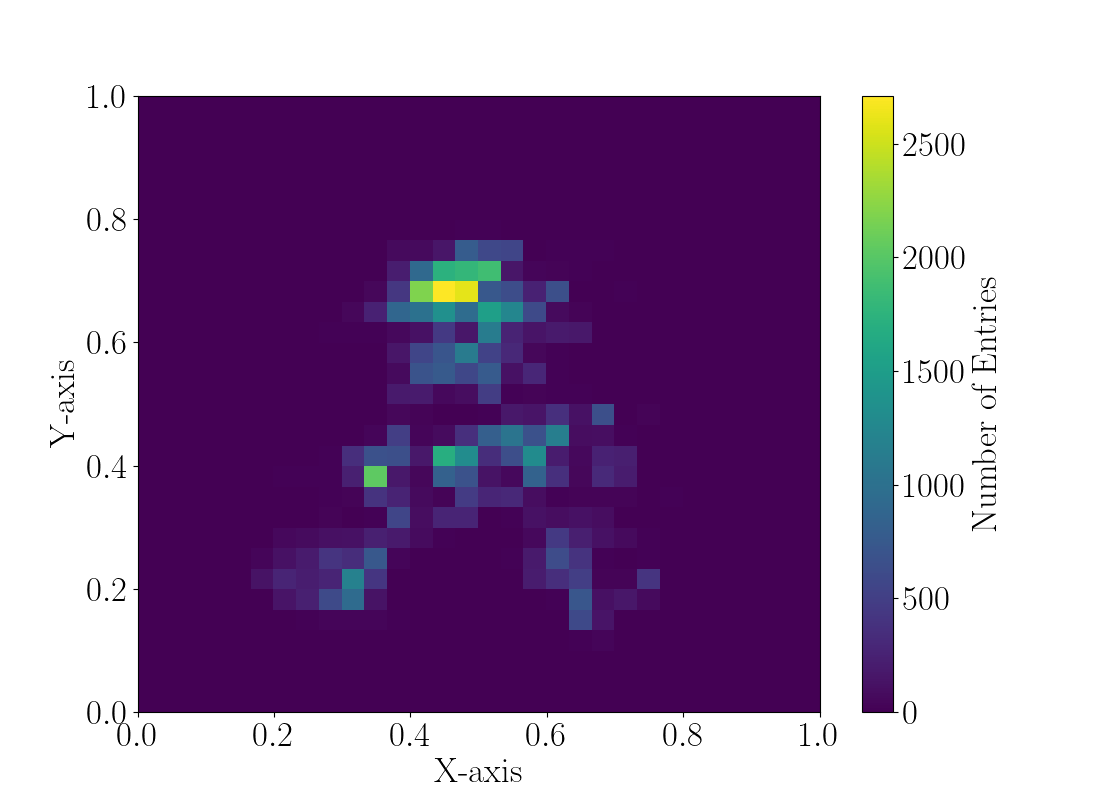}
    }
    \caption{
    (a) The 2D histogram in the case of high-concentration is shown.
    (b) The 2D histogram in the case of low-concentration is shown.
    }
    \label{fig:kpg}
\end{figure}

The motive of this paper is the fact that a human's body movements can be a factor in recognizing his/her condition.
We propose that the standard deviations of designated points are a core factor in measuring the concentration levels.
To measure the points, we used OpenPose \cite{b6, b7, b8, b9}, which is shown in Fig. \ref{fig:wdlee} with the coordinate data of the points range from 0 to 1.
We then check the distributions when humans are concentrating or not respectively, which is shown in Fig. \ref{fig:kpg}.
The distribution in (a) shows that the entries are gathered around body points while (b) is more widely spread.
The difference is distinguished, but it cannot be quantified to recognize the concentration levels.
We analyze the distributions by using DNN-K, and the method details are discussed in the following sections.

\section{Methods} \label{methods}
\subsection{Overview}
\begin{figure}[h!]
    \centering
    \includegraphics[width=0.7\columnwidth]{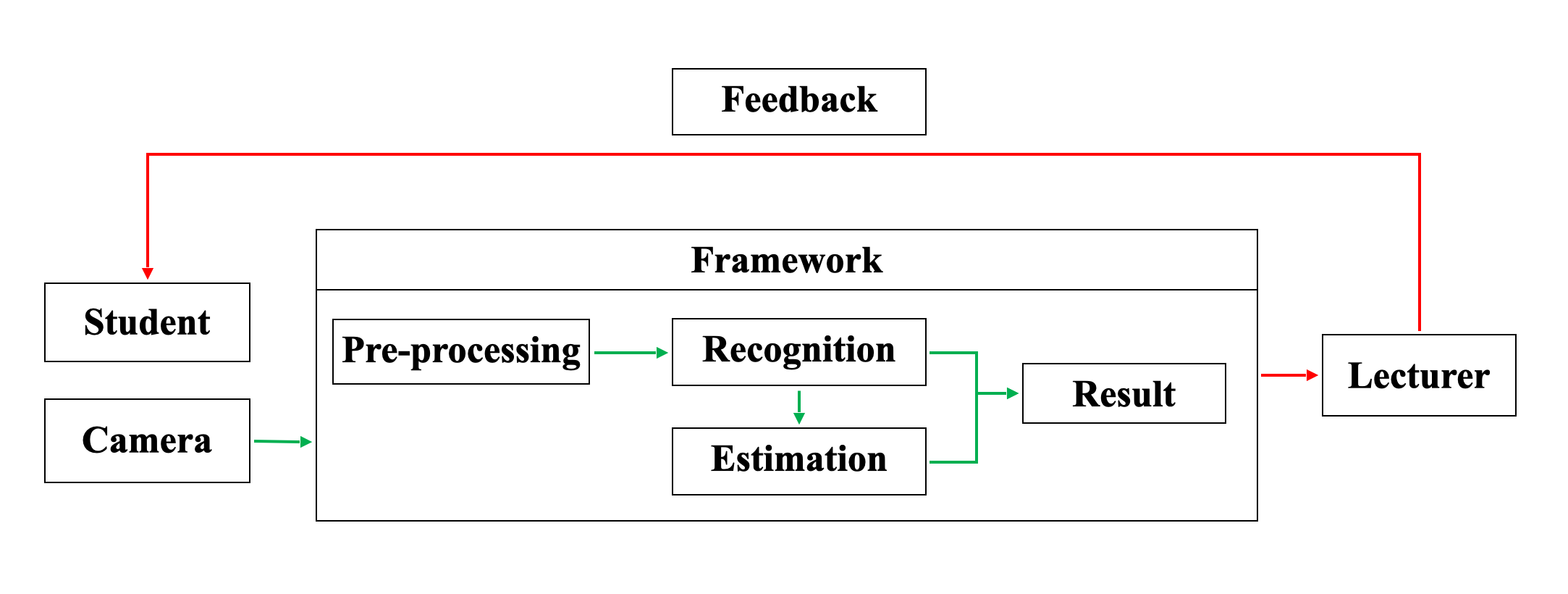}
    \caption{The overview of our research is shown.
    DNN-K consist of several packages to recognize and estimate the levels.
    }
    \label{fig:diagram}
\end{figure}
Figure \ref{fig:diagram} shows the overview of DNN-K.
The first step of DNN-K is where the student's video data taken by a camera is pre-processed.
The pre-processed data is labeled by two states, high and low concentration levels, based on the students' intent.
The labeled data is used for training a DNN model at the recognition step.
The trained DNN model recognizes the continuous concentration levels, which is defined as recognition levels ($S_{r}$). 
The concentration levels are estimated by KF in the estimation step, which are called the estimation levels ($S_{e}$).
At the end, lecturers can give their feedback to students with $S_{r}$ and $S_{e}$.

\subsection{Step 1. Data pre-processing} 

The first step of DNN-K is to extract the standard deviations from video data.
Ten points of the human body are measured every 50 frames, which are classified as the top part (0 - 4) and the middle part (5 - 9).
Then, the standard deviations of the X and Y coordinates are calculated in the top and the middle parts respectively.
Note that we assume the standard deviations of the points are the core factor in measuring the concentration levels.
Table \ref{tab:dataStructure} shows the notations of the results in the pre-processing step.

\begin{table}[h!]
    \centering
    \begin{tabular}{|c||c|}
         \hline 
         Columns & Description \\
         \hline
         $\sigma_{Top}^X$ & The standard deviation of the top part's X coordinate\\
         $\sigma_{Top}^Y$ & The standard deviation of the top part's Y coordinate\\
         $\sigma_{Mid}^X$ & The standard deviation of the middle part's X coordinate\\
         $\sigma_{Mid}^Y$ & The standard deviation of the middle part's Y coordinate\\
         \hline
    \end{tabular}
     \caption{The definition of the data and its description are shown.
     }
    \label{tab:dataStructure}
\end{table}

Algorithm \ref{algo:dataPre} shows the entire process of the pre-processing step.
%%%%%%% Description of Algorithm %%%%%%%%%
Through this algorithm, the standard deviations are obtained, and become the input data for DNN, discussed in the next section.

\begin{algorithm}
\caption{Data Pre-Processing Step of DNN-K}
\label{algo:dataPre}
\renewcommand{\algorithmicrequire}{\textbf{Input:}}
\renewcommand{\algorithmicensure}{\textbf{Output:}}
\renewcommand{\algorithmicforall}{\textbf{for each}}

\begin{algorithmic}
% \REQUIRE Video data
\REQUIRE top.X, top.Y, mid.X, mid.Y
\begin{comment}
    \STATE /* Detecting points */
    \FOR{\(frame \in video\)}
        \FOR{\(point \in body\)}
            \STATE \(point.X, point.Y \leftarrow Open Pose (frame)\)
        \ENDFOR
    \ENDFOR
    \STATE /* Classifying data */
    \FOR{\(point \in body\)}
        \IF{\(point \in Top\)} \(Top.X, Top.Y \leftarrow point.X, point.Y\) \ENDIF
        \IF{\(point \in Mid\)} \(Mid.X, Mid.Y \leftarrow point.X, point.Y\) \ENDIF
    \ENDFOR
\end{comment}
\FORALL{\(Data \in \{top.X, top.Y, mid.X, mid.Y\}\)}
    \FOR{\(D_i = \underset{50i}{\overset{50(i+1)}{\bigcup}}Data\)}
        \STATE \(\sigma_{Data} \text.append({s.d.}(D_i))\)
    \ENDFOR
\ENDFOR

\ENSURE \(\sigma_{Top}^X, \sigma_{Top}^Y, \sigma_{Mid}^X, \sigma_{Mid}^Y\)
\end{algorithmic}
\end{algorithm}

\begin{figure}
    \centering
    \subfigure[]{
        \includegraphics[width=0.35\columnwidth]{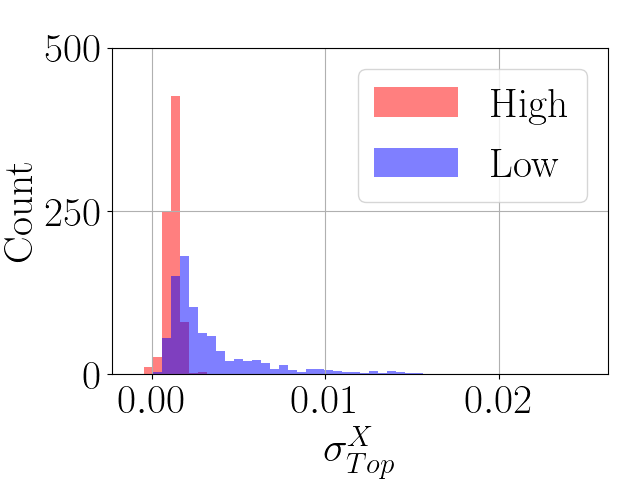}
        \label{fig:topX}
    }
    \subfigure[]{
        \includegraphics[width=0.35\columnwidth]{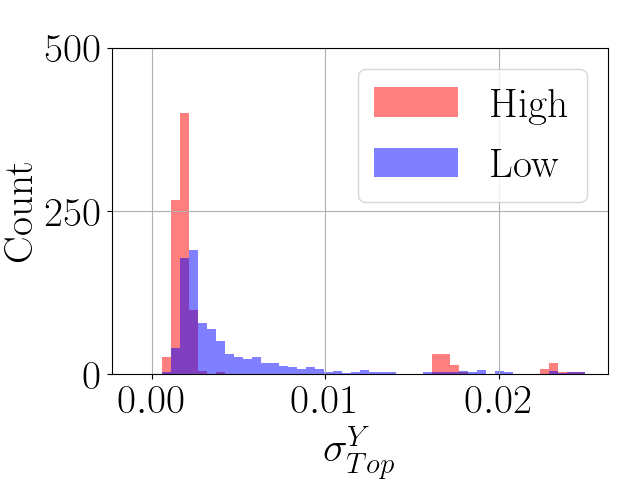}
        \label{fig:topY}
    }
    \subfigure[]{
        \includegraphics[width=0.35\columnwidth]{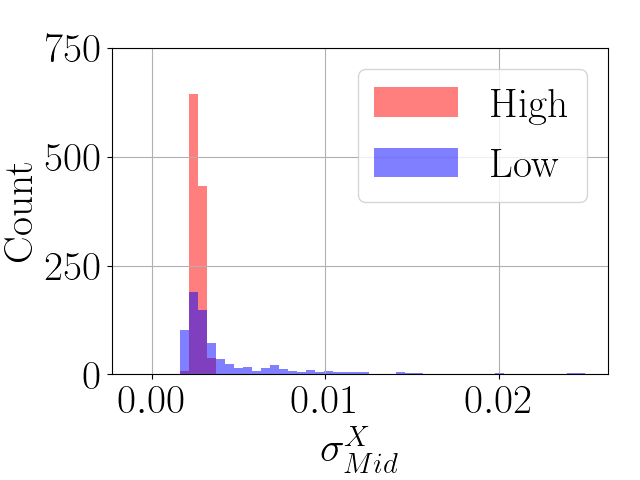}
        \label{fig:midX}
    }
    \subfigure[]{
        \includegraphics[width=0.35\columnwidth]{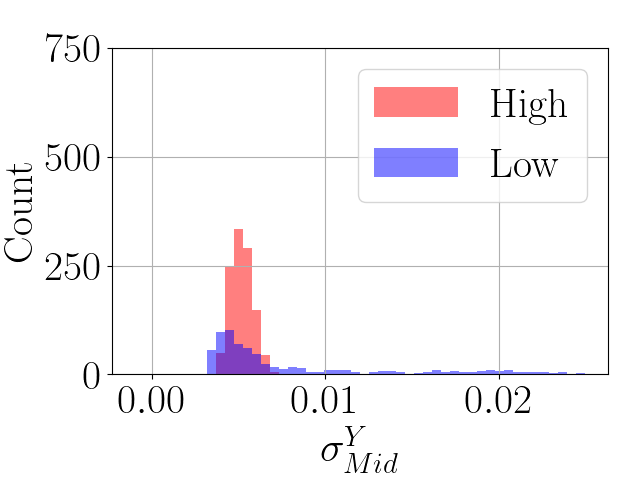}
        \label{fig:midY}
    }
    \caption{
    (a) - (d) show the distribution of the standard distribution for each parts, respectively. 
    Red histogram for high concentration case, and blue histogram for low concentration case are shown.
    }
    \label{fig:topmid}
\end{figure}
Figure \ref{fig:topmid} shows the difference of the standard deviations among each group.
Nevertheless, there remain unexplained aspects, such as ambiguous patterns, the correlation of which with the concentration levels are not clear.
To this end, DNN is applied to solve the problems as DNN is an appropriate method for obtaining nonlinear combinations from features.
This allows us to unveil hidden features that we cannot acknowledge.

\subsection{Step 2. Recognition} \label{step2}

\begin{algorithm}
\caption{Recognition Step of DNN-K}
\renewcommand{\algorithmicrequire}{\textbf{Input:}}
\renewcommand{\algorithmicensure}{\textbf{Output:}}

\begin{algorithmic}
\REQUIRE $\sigma_{Top}^X, \sigma_{Top}^Y, \sigma_{Mid}^X, \sigma_{Mid}^Y$

\STATE \(\text{Input layer  :} \in R^4\)
\STATE \(1^{st}\text{ hidden layer :} \in R^8 \text{ (activation function: ReLU)}\)
\STATE \(2^{nd}\text{ hidden layer :} \in R^8 \text{ (activation function: ReLU)}\)
\STATE \(\text{Output layer :} \in R^1 \text{(activation function: Sigmoid)}\)

\ENSURE $\textsl{Concentration Levels}$ $s_{t}$ $\in$ S
\end{algorithmic}
\end{algorithm}
Algorithm 2 shows the overall structure of the recognition step. 
Through our DNN, the recognition levels ($S_{r}$) are obtained.
The DNN consists of four layers.
ReLU is used as the activation function in the first hidden layer and the second hidden layer.
Sigmoid is used to make sure the probability is distributed relatively evenly from 0 to 1. 
ADAptive Moment (ADAM) estimation optimizer \cite{b10} is applied and the initial learning rate is set as 0.1 $\%$, which is the optimal value for the DNN.
Loss function ($L$) is defined as 
\begin{align}
  L = -\frac{1}{N}\sum_{i=1}^{N}[y_{i}\log(\hat{y_{i}})+(1-y_{i})\log(1-\hat{y_{i}})], 
  \label{eq:binary}
\end{align}
where N is the number of labels, which is two in this case. $y_{i}$ is the label of the data and $\hat{y_{i}}$ is the probability of the prediction value from the data as $y_{i}$.
As the output value is a probability for binary classification, we use binary cross-entropy for calculating the loss for every epoch during training.

K-Fold is applied to cover the insufficient data.
The accuracy of 5-Fold training is ranged from $85 \%$ to $95 \%$ with a median of 90.62 $\%$.

\subsection{Step 3. Estimation} \label{step3}

\begin{algorithm}
\caption{Estimation Step of DNN-K}
\renewcommand{\algorithmicrequire}{\textbf{Input:}}
\renewcommand{\algorithmicensure}{\textbf{Output:}}

\begin{algorithmic}
\REQUIRE $s_{t} \in S$
\FORALL{$s_{t} \in S$}
    \STATE \(\bold{x_{t}^{est}} \leftarrow {s_{t}}\)
    \STATE /* Predicting */
    \STATE \(\bold{x_{t+1}^{pre}} = \bold{A} \cdot  \bold{x_{t}^{est}}\)
    \STATE \(\bold{P_{t}^{pre}} = \bold{A} \cdot \bold{P_{t}} \cdot \bold{A^{T}} + \bold{Q}\)
    
    \STATE /* Updating */
    \STATE \(\bold{K} = \bold{P_{t}^{pre}} \cdot \bold{H}/ (\bold{H} \cdot \bold{P_{t}^{pre}} \cdot \bold{H} + \bold{R})\)
    
    \STATE /* Estimating */
    \STATE \(\bold{x^{est}_{t+1}} = \bold{x_{t+1}^{pre}} + \bold{K} \cdot (\bold{m_{t}} - \bold{H} \cdot \bold{x_{t+1}^{pre}})\)
    \STATE \(\bold{P_{t+1}} = \bold{P^{pred}_{t}} - \bold{K} \cdot \bold{H} \cdot \bold{P^{pred}_{t}}\)
\ENDFOR
%%%%%%%
\STATE /* Analyzing */
\STATE \(\text{Fitting the distribution of } \bold{x^{est}_{t+1}} \text{ by lecturers'}\)
\STATE \(\text{defined function}\) 
\ENSURE $\textsl{Concentration Levels}$ $(\Psi)$
\end{algorithmic}
\end{algorithm}
The estimation step of DNN-K includes KF to earn $S_e$.
Algorithm 3 shows the overall process.
There are three states in the algorithm, which are described as prediction state ($\bold{x_t^{pre}}$), estimation state ($\bold{x_t^{est}}$), measurement state ($\bold{m_t}$), the error covariance matrix ($\bold{P_t}$), and the transition weight matrix ($\bold{A}$). 
The students' state starts from $\bold{x_0^{est}} = 0.5$,  because the students' concentration level is assumed as $50 \%$ in the beginning.
$\bold{P_0} = 0.9$ is the system error, which comes from the DNN, discussed in the previous section.

In the predicting step, $\bold{A}$ is a transition matrix, and $\bold{Q}$ is an external noise matrix, which can be modified by teachers.
$\bold{A}$ is set to $\mathbb{1}$ and $\bold{Q}$ is set to $\mathbb{0}$ as an ideal case.
The students maintained their concentration levels and there were no external disturbances when they took lectures. 
In the updating step, the Kalman Gain ($\bold{K}$) is obtained in every step.
$\bold{H}$ is a scale matrix, which is set to $\mathbb{1}$ by simplifying the problems.
$\bold{x_{t+1}^{est}}$ and $\bold{P_{t+1}}$ are updated with $\bold{K}$. 
Finally, in the estimating step, the next estimated state $\bold{x_{t+1}^{est}}$ is recurrently updated.

\begin{figure}
    \centering
    \includegraphics[width=0.7\columnwidth]{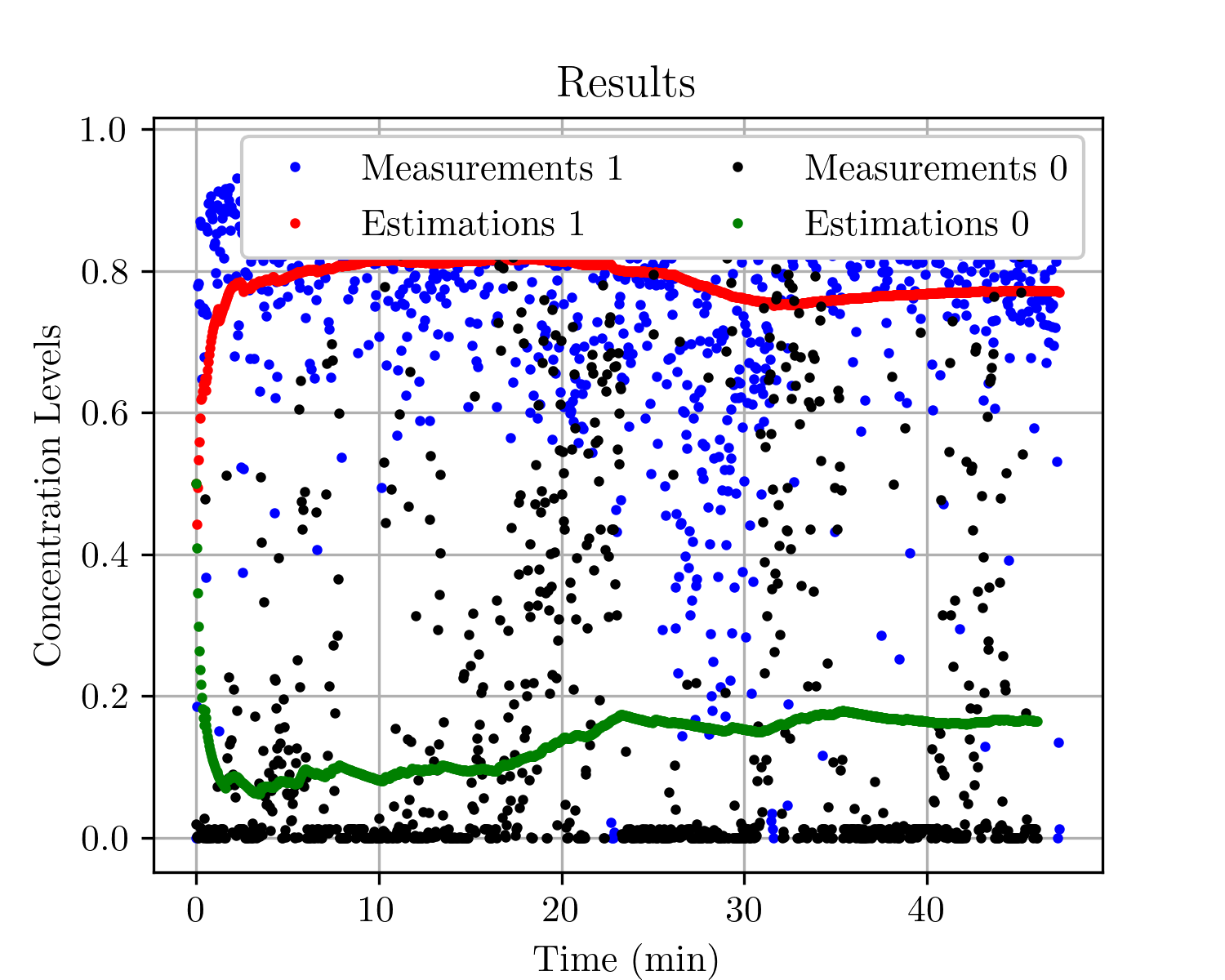}   
    \caption{
    Estimation and measurement levels are shown.
    % This figure shows an example of the estimation result of concentration level with a simple KF. 
    % Every 2.5 seconds which are shown as Estimations 0 and Estimations 1 with red and green dots.
    The black dots ($S_r$) and the green dots ($S_e$) show a low concentration case. 
    The blue dots ($S_r$) and the red dots ($S_e$) show a high concentration case. 
    }
    \label{fig:resultKF}
\end{figure}

\begin{figure}
    \centering
    \includegraphics[width=0.7\columnwidth]{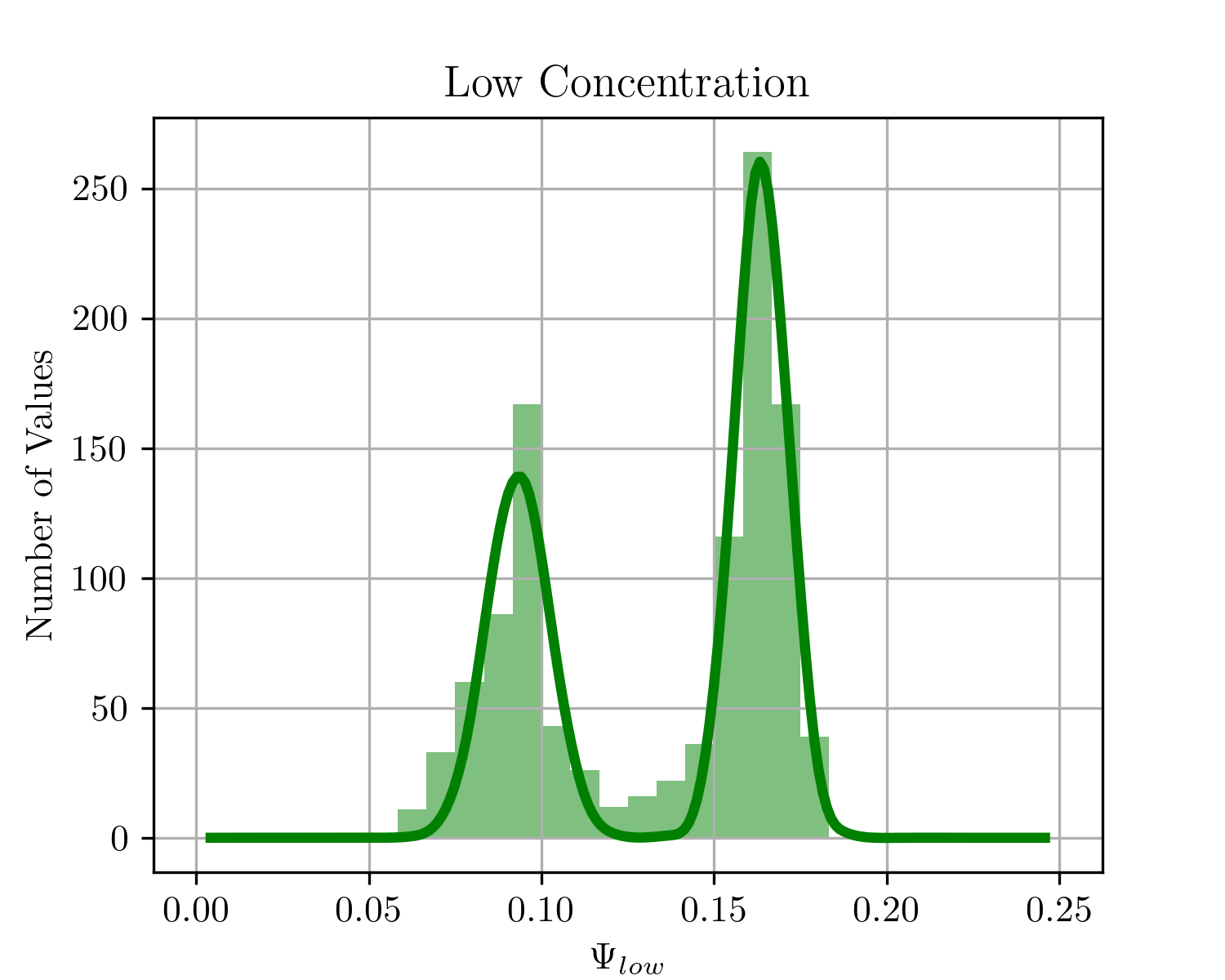}
    \caption{Fitting result in the low-concentration case is shown. 
    Two separated states could be described as Eq. (\ref{eq:fitting}).}
     \label{fig:resultLow}
\end{figure}

Figure \ref{fig:resultKF} shows the estimation and measurement results every 2.5 seconds.
Even though the measurement fluctuates widely every 2.5 seconds, KF enables users to track the levels smoothly, which are shown as the green and the red dots.
Figure \ref{fig:resultLow} shows the histogram of the green dots.
In this case, the distribution consist of two dominant modes,  which can be analyzed by a certain function, and the details are as follows.
Each mode can be described by the function of bimodal distribution $X$, which is written as
\begin{ceqn}
\begin{align}\label{eq:fitting}
     N_{k}(\mu_{k}, \sigma_{k})  &= A e^{-\frac{(x-\mu_{k})^2}{2\sigma_{k}^2}}\\
    X &= \mathcal{N}_{1}(\mu_1,\,\sigma_1^{2}, A_1)\ + \mathcal{N}_{2}(\mu_2,\,\sigma_2^{2}, A_2)\,
\end{align}
\end{ceqn}
where $\sigma_1$ and $\sigma_2$ are the standard deviations, $\mu_1$ and $\mu_2$ are the mean values, and $x$ is the input data.

Figure \ref{fig:resultLow} shows the distribution of $\Psi$ in the low concentration case ($\Psi_{low}$).
The $\mu_1$ and $\mu_2$ are obtained as 0.09 and 0.16 respectively in this distribution.

\section{Conclusion and future work}
We devise the model for aiding lecturers to estimate students' concentration level using cameras in online classes. 
Our system presents the level every 2.5 seconds under 90.62 $\%$ accuracy and estimates the next level of concentration by using KF.
In contrast to the previous research, such as using VGG16 \cite{b11}, our model takes a different approach to quantify the levels by captivating the variance of detected points on humans in the current state.
Additionally, we estimate and track the level for the next time window.
Our model practically offers a tool to monitor the level more precisely and aid lecturers to estimate the level.
Academically, our model has a novel approach to analyze complex human states and the concentration level. 

As a future work, we plan to use not solely body movement data but also emotion data \cite{b12} and skin thermal data \cite{b1} \cite{b13} for an enhanced prediction of measuring human concentration levels.
The measuring method used in this paper and the conventional measuring techniques will be combined and processed using deep learning.
This work expects to provide useful information on students' concentration level and thus assist lecturers.

% \end{multicols}

\end{document}